\documentclass{llncs}
\usepackage[utf8]{inputenc} 
\usepackage{epsfig}
\usepackage{amsmath}
\usepackage[colorinlistoftodos]{todonotes}

\begin{document}

\title{Leishmaniasis Parasite Segmentation and Classification using Deep Learning}
\titlerunning{Leishmaniasis Parasite Segmentation and Classification using Deep Learning}  
%
\author{Marc Górriz\inst{1}, Albert Aparicio\inst{1}, ,
Berta Raventós\inst{2}, Verónica Vilaplana\inst{1}, Elisa Sayrol\inst{1} and Daniel López-Codina \inst{2}  \\
}
\authorrunning{Marc Górriz et al.} 
%
%


\institute{Image and Video Processing Group, Signal Theory and Communications Department\\
Universitat Politecnica de Catalunya (UPC)\\
Barcelona, Catalonia/Spain\\
{\tt\small \{veronica.vilaplana, elisa.sayrol\}@upc.edu}
\and
Computational Biology and Complex Systems Group, Physics Department\\
Universitat Politecnica de Catalunya (UPC)\\
Barcelona, Catalonia/Spain\\
{\tt\small daniel.lopez-codina@upc.edu}
}

\maketitle              

\begin{abstract}
Leishmaniasis is considered a neglected disease that causes thousands of deaths
annually in some tropical and subtropical countries. There are various techniques to diagnose leishmaniasis of which manual microscopy is considered to be the gold standard. There is a need for the development of automatic techniques that are able to detect parasites in a robust and unsupervised manner.
In this paper we present a procedure for automatizing the detection process based on a deep learning approach. We train a U-net model that successfully segments leismania parasites and classifies them into promastigotes, amastigotes and adhered parasites.

\end{abstract}


%
%

\section{Introduction}
Leishmaniasis is a disease that next to malaria,  is the second worst known parasitic killing disease; an estimated 700.000 to 1 million new cases and 20.000 to 30.000 deaths occur each year according to the World Health Organization \cite{WHO}. 

Direct observation of the leishmaniasis parasite in the microscope can be considered the gold standard for diagnosis. Nevertheless, it requires time and technical expertise, and because of the quantity of steps needed for manual diagnosis, this analytic technique is tedious and inclined to human mistake even in experienced hands, leading to possibly late conclusions and mistaken diagnosis.
The purpose of this paper is to propose an alternative to manual observation of the leishmania parasites, as this process is hard and time consuming, by creating a system that automatizes the detection procedure. 

Although automatic analysis of blood cells and other medical images have been under study for the last 50 year, just in the last decade there has been an increasing interest in processing images of intracellular protozoan parasites, including the processing of images acquired from samples of neglected diseases such as malaria, Chagas or leishmaniasis. In particular, automatic detection of leismaniasis through image processing techniques has been addressed by some authors. Considering the standard staining, morphological and computer vision methods to segment parasite bodies have been utilized in \cite{farahi2015cvlevel}. 
Another approach considering a watershed based segmentation technique is presented in \cite{vazquez2013morpho}, where internal and external markers are defined based on the fact that parasites are present in larger numbers than the host cells. In \cite{yazdanparast2014inspect} a devoted software called INsPECT was developed to automate infection level measurement based on fluorescent DNA staining. They also use morphological filtering as a preprocessing step followed by what they call a Threshold for images with Decreasing Probability Density Function. Other methods that also use fluorescent staining and Giemsa staining can be found in the literature \cite{ouertani2014}, \cite{ouertani2016}.

In the last years deep learning techniques have shown a disruptive performance in different image processing and computer vision applications. From the seminal work of \cite{krizhevsky} for image classification, many new architectures and applications have been addressed. In particular, convolutional neural networks (CNN) have rapidly become a methodology of choice for biomedical image analysis. Recently, some systems based on CNNs have been proposed for microscope-based diagnosis in resource-constrained environments, using photographs of samples viewed from the microscope. For example, \cite{Quinn} evaluates the performance of deep convolutional neural networks on three different microscopy tasks: diagnosis of malaria in thick blood smears, tuberculosis in sputum samples, and intestinal parasite eggs in stool samples. In all cases accuracy is high and substantially better than alternative approaches based on traditional medical imaging techniques. \cite{Mehnian} proposes a computer vision system that leverages deep learning to identify malaria parasites in micrographs of standard, field-prepared thick blood films. They train a CNN based on a VGG architecture for feature extraction and use logistic regression for classification. \cite{penas} also use a CNN to detect and identify some species of malaria parasites through images of thin blood smears. To the best of our knowledge, however, there are no previous works using deep learning for leishmania parasite segmentation and classification on microscopy images.

In this paper we present an automated system based on a fully convolutional neural network for segmenting leishmania parasites and classifying them into promastigotes, amastigotes and adhered parasites. 

Following in Section 2 we describe the database that was used in this work. Section 3 is devoted to the U-Net and the method that has been developed to detect parasites. Results are presented in Section 4 and Conclusions are finally exposed in Section 5.


%

%
%
\section{Data}

Database was provided by the Computational Biology and Complex Systems Group at Universitat Politècnica de Catalunya. Cultures were generated from macrophage infection of RAW cells 264.7 with Leishmania infantum, Leishmania major and Leishmania  braziliensis and observed after 48 hours. Images were obtained with the microscope using the light transmission and the brightfield technology widely used in biological preparations previously fixed and Giemsa staining.  Preparations were illuminated from the bottom of the equipment. Properties were not altered (devices such as polarizers or filters were not used). Images were captured with a magnification of 50x to facilitate image analysis.

\begin{figure}[t]
\begin{center}
\includegraphics[width=0.9\linewidth]{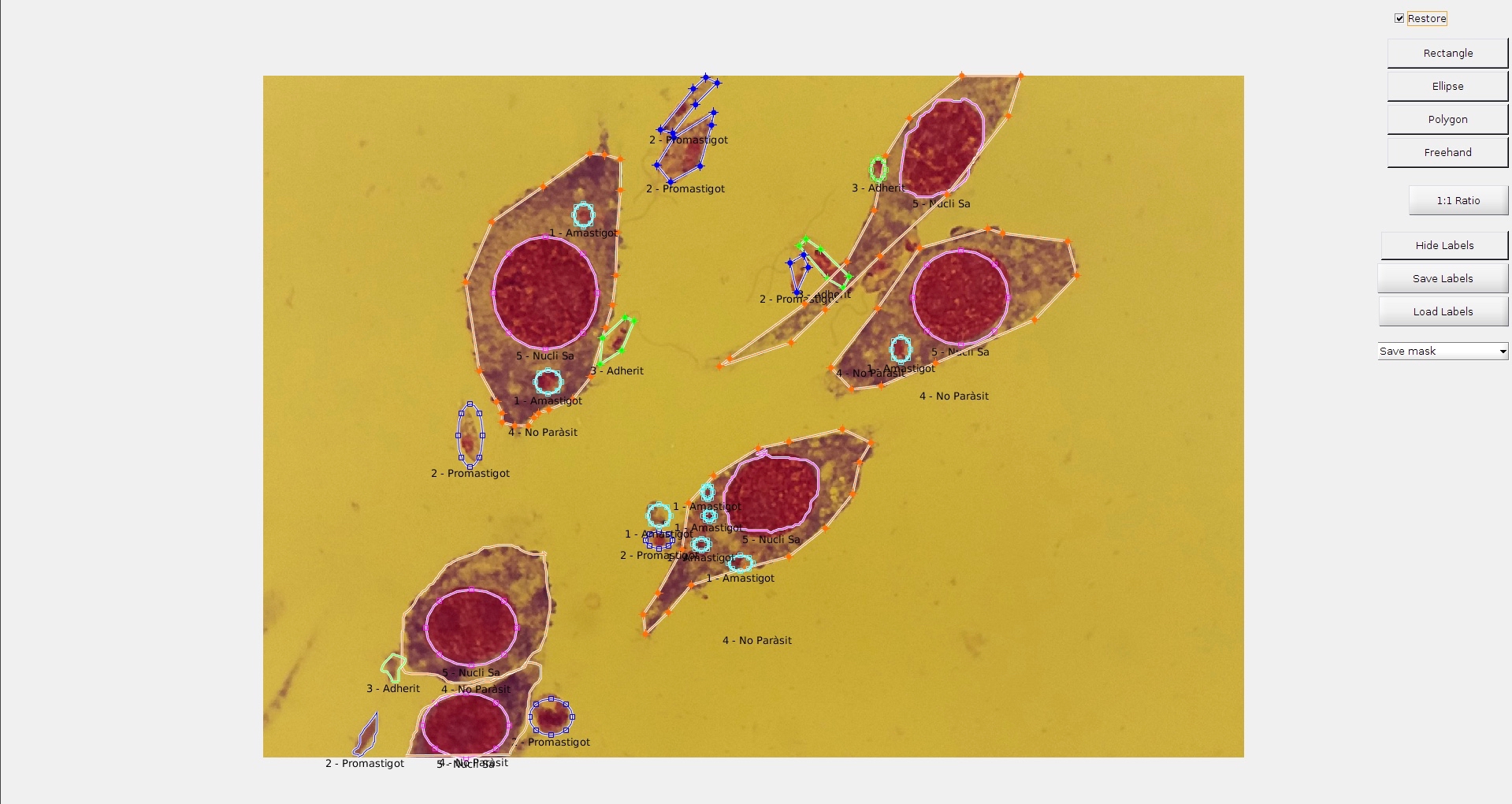}
\end{center}
\caption{Annotation Tool with different labeled regions.}
\label{fig:annotated_image}
\end{figure}

The number of images provided was 45, sizes are around 1500x1300pixels.  Images show promastigote shapes of the parasite when cells have not yet been infected. They have a fusiform aspect. 
Amastigote shapes of the parasite appear once cells have been infected and show a very simple structure in stained images. They are seen as small oval corpuscles.
There is an interest too in distinguishing those promastigotes that are just adhered to the cell membrane. The different elements of images can be observed in Figures \ref{fig:annotated_image} and \ref{fig:results}. 

To train the automatic detector with ground truth examples, images were annotated with the supervision of an expert using a tool specifically developed for this project. An example can be seen in Figure \ref{fig:annotated_image}. Image regions belonging to promastigotes, amastigotes and adhered shapes were annotated. Also cell cytoplasm membrane and its nucleus were labeled. Other regions that were not parasites were annoted as "unknown" they were caused by stain blobs or other artifacts. Regions not labeled are considered background.


\section{Method}

\begin{figure}[t]
\begin{center}
\includegraphics[width=0.95\linewidth]{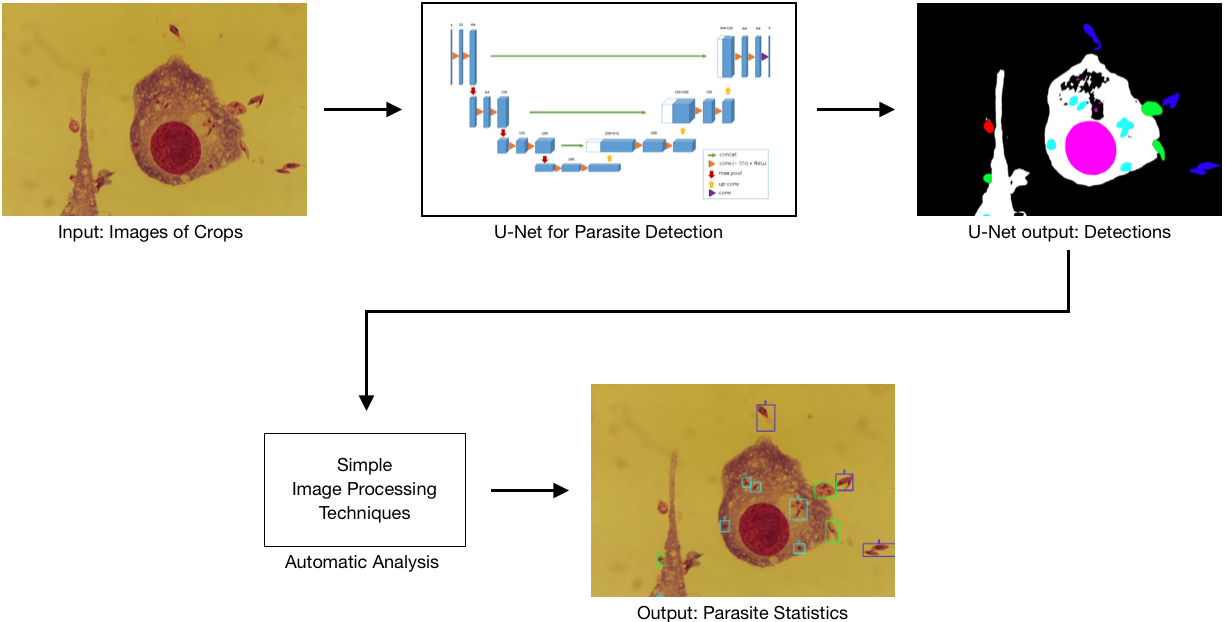}
\end{center}
\caption{Method pipeline for detecting Leishmaniasis parasites.}
\label{fig:method_pipeline}
\end{figure}

This section introduces the methodology used for detecting Leishmania parasites. The first step is the segmentation of the input image, where each pixel is individually classified using a fully convolutional CNN. Then, a post-processing stage estimates the number and size of each parasite type (promastigote, amastigote and adhered parasite). Figure \ref{fig:method_pipeline} shows the complete pipeline.

\subsubsection{U-Net architecture.}

\begin{figure}[t]
\begin{center}
\includegraphics[width=0.95\linewidth]{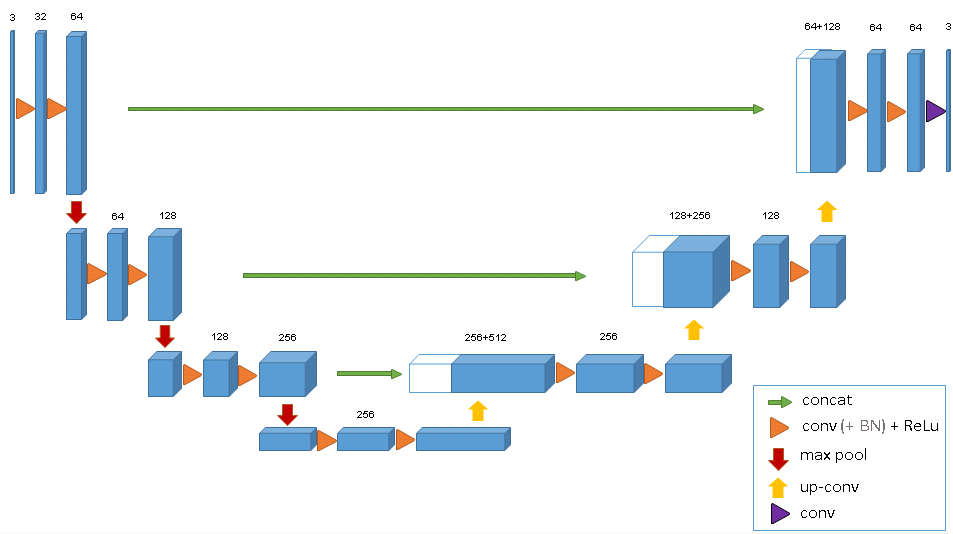}
\end{center}
\caption{U-Net architecture (example for RGB input samples with multi-class output).} 
\label{fig:unet}
\end{figure}

We use U-Net, a fully convolutional network model proposed in \cite{unet} (Figure \ref{fig:unet}). The network combines a convolutional network (a contracting path) with a deconvolutional network (an expanding path).   
The network consists of the repeated application of two $3$x$3$ convolutions, followed by non-linear activations (ReLU) and a downsampling process through $2$x$2$ maxpooling  with stride $2$. 
The use of pooling layers in deep networks provide coarse, contextual features. In order to obtain fine details in the segmentation map, the U-Net combines multi-scale features by connecting corresponding resolutions in the contracting and the expanding path.
The deconvolutional path consists of an upsampling of the feature maps followed by a $2$x$2$ convolution (halving the number of filters), concatenation of the corresponding feature maps from the contracting path, and two $3$x$3$ convolutions followed by ReLU activations. The entire network has $23$ convolutional layers. The last layer is used to map each component feature vector to the desired number of classes in order to generate the output pixel-wise segmentation map. 

\subsubsection{Training.}
The training set consists of 37 images, each one with its corresponding ground truth
image containing labels for seven classes: promastigote, amastigote, adhered parasite, nucleus, cytoplasm, background and unknown.

We train the network end to end with patches of size [224 x 224] extracted over the training images with an overlap of 112 pixels.

The dataset is characterized by a very high class imbalance, since most image pixels correspond to background and only a small percentage of pixels correspond to parasite classes (see Table \ref{tab: class-balance}).
High class imbalance may drive the network to predict the most common class in the training set. In order to alleviate this problem, we adopt two strategies. We apply a two-stage non-uniform sampling scheme to select the training patches, and we use as loss function the Generalized Dice Loss \cite{sudre}.


\begin{table}
\caption{Class distribution of the Leishmania dataset.}
\begin{center}
\begin{tabular}{|c||c|c|c|c|c|c|c|}
\hline
\textbf{Class}  & Background & Cytoplasm & Nucleus & Promastigote & Adhered & Amastigote & Unknown \\
\textbf{Pixels} & 99.38 \%      & 0.41 \%     & 0.16 \%    & 0.02 \%      & 0.01 \%  & 0.02 \%      & 0.59 \% \\ \hline
\end{tabular}
\end{center}
\label{tab: class-balance}
\end{table}

Regarding the sampling strategy, the network is first trained during some epochs (40 epochs in the experiments presented in Section \ref{sec:results}) using patches that contain at least 40 \% of pixels from any of the three parasite classes (promastigote, amastigote, or adhered), and then uniform sampling of all patches is used for the following epochs (200 in the experiments). 

Data augmentation is used to increase the amount of training data by applying random transformations such as rotations, horizontal and vertical flips and their combinations.


The loss function used is the Generalized Dice Loss, proposed in \cite{sudre} to mitigate the class imbalance problem. For ground truth pixel values $r_{ln}$ and predictions $p_{ln}$ it can be expressed as:
\begin{equation}
  \mathcal{L} = 1 - 2 \frac{\sum_{l=1}^C w_l \sum_{n} r_{ln}p_{ln}}{\sum_{l=1}^C w_l \sum_{n} r_{ln} + p_{ln}}
\end{equation}
where $C$ is the number of classes and $w_l$ is used to provide invariance to different label set properties. We set $w_l = 1 / (\sum_{1}^N r_{ln})^2$  to reduce the correlation between region size and Dice score, by correcting the contribution of each label by the inverse of its area. 



\begin{figure}[t]
\begin{center}
\includegraphics[width=0.6\linewidth]{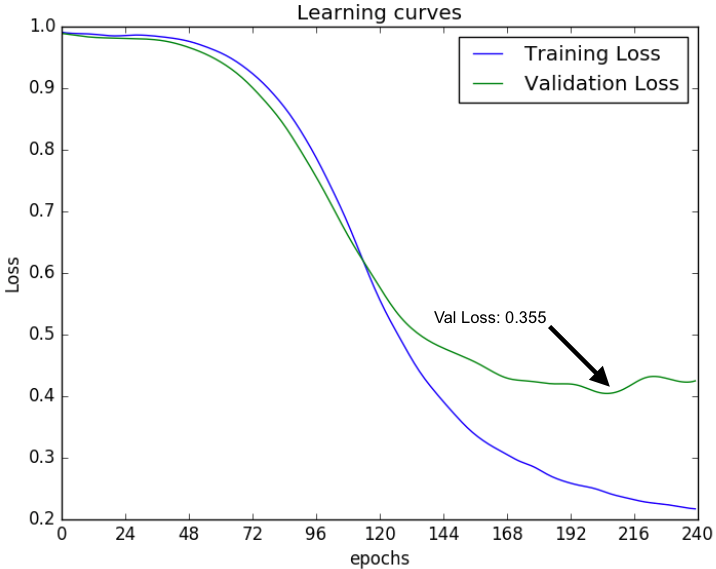}
\end{center}
\caption{Learning curves for U-Net model (Generalized Dice Coefficient Loss).} 
\label{fig:lcurves}
\end{figure}

We used Adam optimizer with a learning rate of $1\mathrm{e}{-4}$. The batch size was $5$. The network took approximately 15 hours to train on a NVIDIA GTX Titan X GPU using the Keras framework with the Tensorflow backend. Figure \ref{fig:lcurves} shows the learning curves.


Once pixels are classified, regions for each of the parasites are defined by considering connected component labeling (CCL). Those regions that are far from the mean size of the class are discarded. This postprocessing allows counting the number of parasites of each class.


%
%
\section{Results}\label{sec:results}


The metrics used for pixel-wise evaluation are: 

\begin{itemize}
\item \textbf{Dice score}: this metric is computed by comparing the pixel-wise agreement between the ground truth (Y) and its corresponding predicted segmentation (X).
\begin{equation}
  Dice \ score = \frac{2 * |X \cap Y|}{|X| + |Y|}
\end{equation}
\item \textbf{Precision},  the ratio of correctly predicted positive observations to the total predicted positive observations. It is defined in terms of True Positives (TP) and False Positives (FP).
\begin{equation}
  Precision = \frac{TP}{TP + FP}
\end{equation}
\item \textbf{Recall}, the ratio of correctly predicted positive observations to the all observations in actual class. It is defined in terms of True Positives (TP) and False Negatives (FN).
\begin{equation}
  Recall = \frac{TP}{TP + FN}
\end{equation}
\item \textbf{F1-score}, the weighted average of Precision and Recall. Therefore, this score takes both false positives and false negatives into account.
\begin{equation}
  F1 \ score = \frac{2 * Precision * Recall}{Precision + Recall}
\end{equation}
\end{itemize}

\begin{table}
\caption{Evaluation of pixel-wise classification in terms of Dice score, precision, recall and F1-score and the percentage of pixels per class.}
\begin{center}
\begin{tabular}{cccccc}
\hline
Class        & Dice score & Precision  & Recall   & F1-score   & Pixels \\ \hline \hline
Background   & 0.981  & 0.983   & 0.978  & 0.980    & 97.07 \%   \\
Cytoplasm    & 0.896  & 0.882    & 0.912  & 0.896    &  1.96 \%  \\
Nucleus      & 0.950  & 0.938    & 0.964  & 0.950    & 0.79 \%  \\
Promastigote & 0.495  & 0.512    & 0.476  & 0.491    & 0.07 \%  \\
Adhered      & 0.707  & 0.677    & 0.379 & 0.457    & 0.05 \%  \\
Amastigote   & 0.777  & 0.757    & 0.823  & 0.777    & 0.06 \%\\ \hline
\end{tabular}
\end{center}
\label{tab: ev-seg}
\end{table}


We used the Jacard index to evaluate the automatic detection of parasitic regions. We used CCL to extract connected component regions over the pixel-wise segmentation maps, for parasite classes (promastigote, amastigote and adhered). Table \ref{tab: ev-obj} shows the percentage of detected regions for each class with Jacard Index (J) values greater than 0.25, 0.5 and 0.8 along with the mean and the standard deviation of J. Being Y the region ground truth and X the predicted pixels, Jacard index is defined as: 
\begin{equation}
  Jacard \ index = \frac{|X \cap Y|}{|X| + |Y| - |X \cap Y|}
\end{equation}

\begin{table}
\caption{Automatic detection evaluation for each class based on the Jacard Index (J).}
\begin{center}
\begin{tabular}{cccccc}
\hline
Class        & J$\geq$0.25 & J$\geq$0.5   & J$\geq$0.75    & Mean  & Std.Dev \\ \hline \hline
Promastigote & 0.54        & 0.52         & 0.50          & 0.41  & 0.14   \\
Adhered      & 0.82        & 0.17         & 0.12          & 0.47  & 0.03   \\
Amastigote   & 0.88        & 0.86         & 0.55          & 0.68  & 0.06 \\ \hline
\end{tabular}
\end{center}
\label{tab: ev-obj}
\end{table}



\begin{figure}[t]
\begin{center}
\includegraphics[width=1\linewidth]{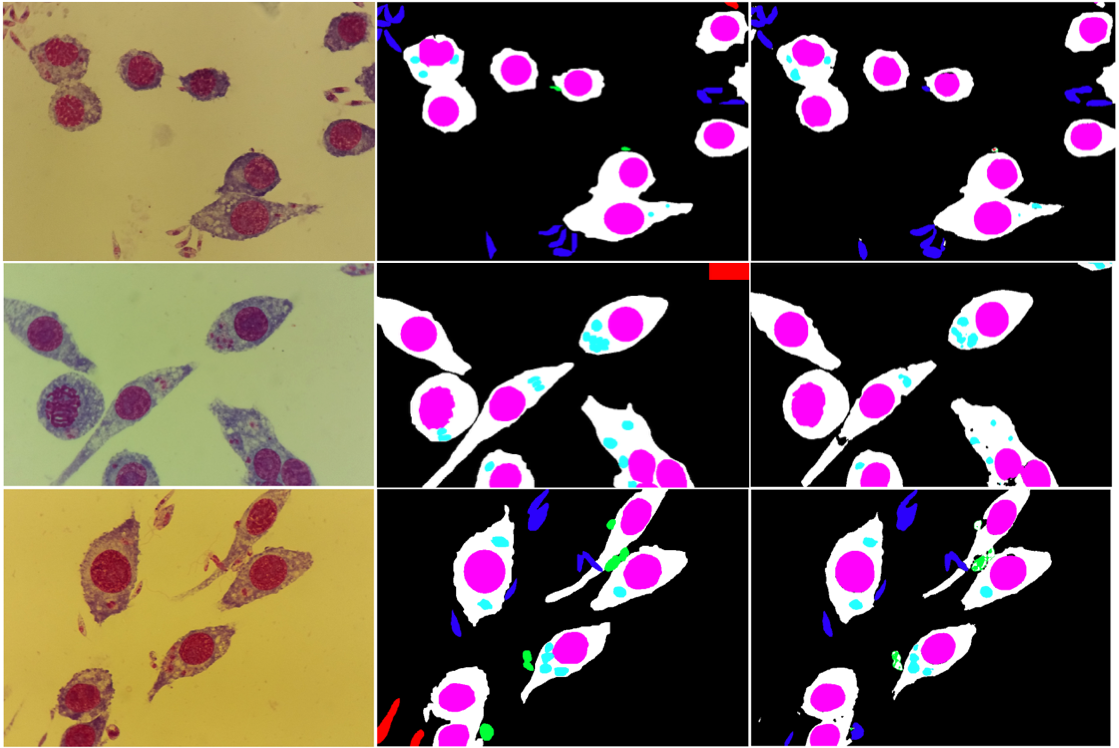}
\end{center}
\caption{Qualitative results for test data. Original images (left), ground truth annotations (middle), network predictions (right).} 
\label{fig:results}
\end{figure}

Observing the results in Table \ref{tab: ev-seg} we appreciate that classes with higher percentage of pixels perform very well, even for the nucleus with a percentage of 0.48\%. Results for the parasite regions with an even lower order of magnitude than nucleus are acceptable. When examining qualitative results in Figure \ref{fig:results} we can observe that ground-truth and predictions for parasite regions look pretty similar. Jacard Indexes in Table \ref{tab: ev-obj} are also acceptable although there is room for improvement. We think that the precision with which annotation was executed can influence the results for such small regions. Thus to raise performance we should consider enlarging the database, and perform a more accurate annotation of the parasite regions. The number of images to train the network was not too high.


%
%
\section{Conclusions}
In this work, we proposed an implementation of a U-Net to detect Leishmaniosis parasites in microscopic images. A training strategy was carried out considering a remarkable class imbalance between background and parasite regions. First by using a non-uniform sampling, that for a given number of epochs trains with patches with a higher percentage of pixels from parasite classes. Second, by choosing the Generalized Dice Loss that mitigates the class imbalance problem. The segmentation maps with test images showed that regions were quite close to ground truth, not only for the cells and its nucleus but also for the three shapes of parasites. Quantitative results using different metrics show promising results that could be improved using larger databases, being the imbalance of the classes and a precise annotation the major drawbacks to cope with.

\section{Acknowledgments}
This work has been developed in the framework of the project TEC2016-75976-R, financed by the Spanish Ministerio de Economía, Industria y Competitividad and the European Regional Development Fund (ERDF).
We gratefully acknowledge the support of the Center for Cooperation and Development to the group of neglected diseases at UPC.
We also give special thanks to Dr. Cristina Riera, Dr. Roser Fisa and Dr. Magdalena Alcover, from the parasitology Section of the Biology, Healthcare and the Environment Department of the Pharmacy Faculty at Universitat de Barcelona advising this work with their knowledge on the Leishmaniosi parasite. We thank the Characterization of Materials Group at UPC to let us use its microscope equipment. Finally we thank Sofia Melissa Limon Jacques for her related work during her Degree Project.
%

{\small
\bibliographystyle{splncs03}
\bibliography{egbib}
}


\end{document}